\documentclass{article}

\usepackage{PRIMEarxiv}

\usepackage[utf8]{inputenc}
\usepackage[T1]{fontenc}
\usepackage{amsmath,amssymb}
\usepackage[round]{natbib}
\usepackage{graphicx}
\usepackage{array}
\usepackage{booktabs}
\usepackage{siunitx}
\usepackage{float}
\usepackage{placeins}
\usepackage{url}
\usepackage[protrusion=true,expansion=false]{microtype}
\usepackage{fancyhdr}
\usepackage[colorlinks=true,linkcolor=blue,citecolor=blue,urlcolor=blue]{hyperref}
\graphicspath{{Figures/}}

\newcommand{\PR}{\operatorname{PR}}
\newcommand{\PRQ}{\ensuremath{\PR_Q}}
\newcommand{\PRK}{\ensuremath{\PR_K}}
\newcommand{\QKt}{\ensuremath{QK^{\top}}}

\title{Query Expansion and Key Specialization in Transformer Attention Geometry
\thanks{Accepted to the main track of the Asian Conference on Machine Learning (ACML 2026).}}

\author{
  Vidit Gupta$^{\dagger}$\quad Siddhesh Nadkarni\quad Mihik Chaudhari\quad Vinaya Sawant\quad Prachi Tawade \\
  \normalfont Dwarkadas J.\ Sanghvi College of Engineering \\
  \normalfont Mumbai, India
}

\begin{document}
\maketitle
{\renewcommand{\thefootnote}{$\dagger$}\footnotetext{Corresponding author: \href{mailto:viditanupgupta@gmail.com}{viditanupgupta@gmail.com}}}

\begin{abstract}
    The projection of queries and keys are central to the attention mechanism in Transformer architectures. While they are mathematically symmetric, they play different roles in attention mechanisms. The question of whether there is an effect from their functional distinction on their geometric development in training remains unanswered. We investigate the problem through the training of small GPT-like Transformers on character-level WikiText-103 for three different depths (4, 6, and 8 layers), three types of initialization for queries and keys, and four random seeds, resulting in 36 runs and 54 trajectories of average layers across seeds. We track the effective dimensionality of those layers using participation ratios and discover that effective dimension of queries expand while keys shrink, and that $PR_Q - PR_K$ is positive in all trajectories studied. In connection to attention, the shrinking of keys leads to a narrower spectrum of $QK^\top$ and more peaked attention weights. In order to determine if this connection is causal or coincidental, we directly control the spectrum of keys during training across five seeds: restricting it to make it shrink sharpens the attention with high directional confidence, while keeping it constant to the level of initial dispersion makes attention softer. Additional token-level checkpoint analyses show that the monotonic paired-contrast trend is not universal across pretrained families, but survives as an early-training regime that later decays over a full pretraining run, and the link between interaction-rank geometry and attention entropy remains visible in several models. 
\end{abstract}


\section{Introduction}

The self-attention mechanism in Transformer-based models is computed using the compatibility matrix \(QK^\top\). Queries and keys are often described as symmetric components of the attention mechanism \cite{vaswani2017attention}. In spite of their mathematical symmetry, the two matrices serve different functional roles: queries represent the information sought by each token position, whereas keys represent the information made available for retrieval. Meanwhile, initialization in deep neural networks is typically viewed as an optimization detail, with Gaussian and Xavier-based schemes primarily used to ensure stable training before the effects of gradient descent dominate \cite{glorot2010understanding}. However, because attention depends directly on the interaction between \(W_Q\) and \(W_K\), the initialization of these matrices may influence the formation of attention patterns during training. Consequently, even when models achieve comparable validation performance, different initialization schemes may lead to distinct attention geometries.

This leads to the main question posed in the paper: do query and key projections follow the same evolution path in terms of effective dimensionality during the training process, or is there a different evolution pattern followed by each one? The answer is sought via participation ratios, spectral-sensitive effective dimensionality measures, wherein \PRQ{} captures how much of the spectral space is being used by query representations, and \PRK{} for key representations

The central result is that there are systematic differences between the geometric paths of query and key projections. The participation of queries increases strongly with respect to depth, seeds, and initialization techniques. The participation of keys diminishes most prominently in the Independent Xavier and Disjoint Support Xavier initializations of $W_Q$ and $W_K$. The contrast between the two slopes, $\PRQ-\PRK$, is positive for the majority of the 216 raw run-layer pairs, and once these are averaged within each seed to remove per-layer noise, the fitted slope is positive for all 54 of the resulting trajectories; this seed-averaged version is the stronger claim reported in the abstract. A raw run-layer pair can still show a negative $\PRQ-\PRK$ value in the earliest epochs, before the slope over the full training window has had a chance to establish itself, which is visible near the start of Figure~\ref{fig:results-participation-summary}. The interventions show that Key-side specialization is closely linked to attention sharpening and selectivity of attention. New checkpoint analyses on token-level models qualify the scope of this statement: the small-model trajectory is not a universal scaling law, but the \QKt{} interaction spectrum remains a useful bridge between key geometry and attention sharpness.

\section{Related Work}

\subsection{Transformer Initialization and Training Stability}

Earlier studies on Transformer initialization focused on the engineering aspect of making it an efficient factor for achieving better convergence and loss stabilization. \citet{glorot2010understanding} established variance-preserving initialization for deep networks, which became the baseline Xavier scheme used in many Transformer implementations. \citet{xiong2020layer} showed theoretically that pre-normalization layer placement removes the depth-dependent gradient explosion of post-normalization architectures, making initialization scale less critical for stability. \citet{liu2020understanding} identified residual branch amplification as a proximate cause of post-normalization instability and proposed Admin initialization to correct it. \citet{huang2020improving} eliminated LayerNorm and warmup through depth-scaled initialization, while \citet{wang2022deepnet} extended stability analysis to very deep Transformers through DeepNorm.

Most of this literature evaluates initialization through optimization stability, convergence speed, or final loss. A broader theoretical perspective views initialization as shaping training-time function-space dynamics \citep{jacot2020neural}. Recent work also suggests that random initialization can induce stable, seed-specific biases that persist throughout training, including SeedPrint-style model identities and structural token-preference biases present at initialization \cite{tong2026seedprints,li2026bornbiased}.

\subsection{Rank Collapse and Entropy Collapse}

A parallel line of work studies geometric degeneration inside Transformers as a pathology to be corrected. \citet{dong2023attention} proved that pure attention without residual connections can contract representational rank rapidly with depth. \citet{noci2022signal} linked initialization-depth interactions to query-key gradient vanishing in residual Transformers. \citet{zhai2023stabilizing} studied attention entropy collapse and proposed spectral reparameterization to prevent unstable concentration. Closest to spectral analyses of attention, \citet{bao2024selfattention} showed that self-attention networks localize when the QK eigenspectrum concentrates, while \citet{hong2025variance} identified softmax variance sensitivity as a driver of attention entropy collapse and training instability.

These studies examine various manifestations of representational collapse, including reductions in rank, concentration of attention, and degeneration of the interaction spectrum.

\subsection{Q/K Subspace Structure and Sparse Attention}

The authors in \citet{pan2024dissecting} analyzed $W_Q^\top W_K$ using singular value decomposition (SVD) in Vision Transformers and identified semantic modes associated with token grouping and contextualization. Previous analyses of attention in pretrained Transformer models reported structured functional roles for attention heads, including syntactic and positional heads, as well as specialized heads that make disproportionately large contributions to model performance \citep{clark2019does, voita2019analyzing}. Other work demonstrated that attention weights alone are insufficient to explain attention outputs because the transformed value vectors and their norms also influence each token's contribution to the resulting representation \citep{kobayashi2020attention}. \citet{tarzanagh2023max} theoretically showed that gradient descent drives the attention softmax toward max-margin token selection. Finally, normalization of Q/K logits directly manipulates the attention-logit geometry, mitigating softmax saturation \citep{henry2020querykey}.
Prior work often analyzes the combined query-key interaction object. Transformer circuit analyses likewise emphasize that the functional QK object is an operator rather than two isolated matrices \cite{elhage2021mathematical}. Induction-head analyses further illustrate how particular attention circuits emerge during training and support sequence-level computations \citep{olsson2022context}. Recent frozen-attention studies provide an additional perspective: \citet{dong2025randomattention} showed that Transformers with frozen query and key weights can still form induction heads and remain competitive on language modeling, while MixiT isolates the role of random fixed attention.

\section{Experimental Setup}

\subsection{Model}

We use decoder-only GPT-style Transformers based on the NanoGPT architecture. Models use 4 attention heads per layer, hidden dimension $d_{model} = 256$, head dimension $d_{head} = 64$, MLP hidden dimension 1024, and context length 128. We evaluate 4-, 6-, and 8-layer variants, yielding approximately 4M--8M parameters depending on depth. All linear projections include bias terms, GELU activations are used in the feed-forward network, and dropout is disabled.

\subsection{Dataset}

We conduct our experiments on the character-level \mbox{WikiText-103} corpus. We use the first 40~million characters of the corpus, split into \SI{90}{\percent} training and \SI{10}{\percent} validation sets. This produces approximately 36\text{M} training tokens, 4\text{M} validation tokens, and a vocabulary size of \num{1758}. Character-level tokenization was chosen to preserve fine-grained sequence statistics and to make early attention dynamics visible in a computationally tractable setting.

\subsection{Training Procedure}

All models were trained using AdamW with learning rate $1 \times 10^{-4}$ , $\beta_1 = 0.9$, $\beta_2 = 0.95$, and weight decay 0.1. Gradient clipping was applied at maximum norm 1.0. We used a constant learning rate schedule, batch size 64, sequence length 128, and mixed precision training with PyTorch AMP.

The baseline observational grid contains 3 depths, 3 initialization regimes, and 4 random seeds (37, 42, 105, and 210), for 36 baseline runs. Across all depths, initialization regimes, and seeds, the observational suite yields 216 run-layer trajectories over a shared 45-epoch analysis window. Unless otherwise stated, baseline statistics are computed over this 36-run grid; when independence is a concern, we report seed-averaged initialization-depth-layer traces rather than treating every layer trajectory as an independent model.

\subsection{Initialization Regimes}

We compare three query-key initialization strategies that explicitly control the geometric relationship between $W_Q$ and $W_K$.

\subsubsection{Independent Xavier}

$W_Q$ and $W_K$ are initialized independently using the standard PyTorch Xavier-uniform initializer, producing near-zero expected alignment between query and key subspaces.

\subsubsection{QK Identical}

$W_Q$ is initialized using Xavier initialization, and $W_K$ is initialized as an exact copy of $W_Q$, producing perfect initial alignment between query and key projections.

\subsubsection{Disjoint Support Xavier}

$W_Q$ and $W_K$ are initialized with non-overlapping support. $W_Q$ contains Xavier-initialized values in the first 128 columns and zeros elsewhere, while $W_K$ contains zeros in the first 128 columns and Xavier-initialized values in the remaining 128 columns. This construction enforces zero initial overlap between active query and key subspaces.

\subsection{Metrics}

To characterize how initialization geometry influences optimization and representation formation, we track metrics spanning optimization behavior, attention concentration, query-key alignment, spectral structure, and representational scaling. Table~\ref{tab:tracked-metrics} lists the tracked quantities. Unless otherwise stated, all metrics are computed layerwise throughout training and aggregated across random seeds.

\begin{table}[!htbp]
\centering
\small
\renewcommand{\arraystretch}{1.15}

\begin{tabular}{@{}>{\raggedright\arraybackslash}p{0.25\linewidth}>{\raggedright\arraybackslash}p{0.71\linewidth}@{}}
\toprule
\textbf{Category} & \textbf{Metrics} \\
\midrule

Optimization &
Training loss, validation loss \\

Attention Distribution &
Attention entropy, top-1/top-5 attention mass, attention gap $p_{(1)} - p_{(2)}$ \\

Query-Key Geometry &
Cosine similarity between $W_Q$ and $W_K$, column-wise alignment \\

Spectral Structure &
Participation ratio (PR), effective rank, covariance eigenvalue statistics \\

Logit Geometry &
Mean, variance, maximum attention logits \\

Norm and Scale &
Activation norms, Frobenius norms, spectral norms \\
\bottomrule
\end{tabular}

\caption{Tracked geometric and optimization metrics.}
\label{tab:tracked-metrics}
\end{table}

The primary geometric metrics used throughout the study are participation ratio (PR) and effective rank:
\begin{equation}
\PR(X)=\frac{\left(\sum_i \lambda_i\right)^2}{\sum_i \lambda_i^2}.
\end{equation}
\begin{equation}
\mathrm{erank}(X)=\exp\left(-\sum_i p_i\log p_i\right),
\qquad
p_i=\frac{\sigma_i}{\sum_j \sigma_j}.
\end{equation}
Here $\lambda_i$ are covariance eigenvalues and $\sigma_i$ are singular values. In this paper, \PRQ{} and \PRK{} are computed from the covariance spectrum of the projected query and key activations for a layer after concatenating heads, so they measure layer-level effective dimensionality rather than per-head semantic feature count. Head-level quantities are used for \QKt{} effective rank and attention effective rank, then averaged across heads for the layer summary. Weight-spectrum diagnostics such as $W_K$ SVD effective rank are computed directly from projection weights and are reported separately from activation PR and \QKt{} interaction-rank metrics.

\subsection{Intervention Experiments}
To test whether query-key geometry plays a causal role in attention organization and spectral evolution, we conducted a suite of controlled interventions on the 8-layer Transformer configuration. These experiments actively modified the dynamics of query and key projections during training, as summarized in Table~\ref{tab:controlled-interventions}.

\begin{table}[!htbp]
\small
\renewcommand{\arraystretch}{1.12}

\begin{tabular}{@{}>{\raggedright\arraybackslash}p{0.22\columnwidth}>{\raggedright\arraybackslash}p{0.74\columnwidth}@{}}
\toprule
\textbf{Intervention} &
\textbf{Manipulation} \\
\midrule

K Spectrum Clamp &
Restores the singular-value spectrum of $W_K$ to its initialization spectrum every 100 optimizer steps from epoch 1 onward \\

Forced K Contraction &
Shrinks the $W_K$ spectral tail after rank 32 by a factor of 0.05 every 100 optimizer steps, then renormalizes the projection \\

Freeze Q &
Restores $W_Q$ to its initialization reference every optimizer step \\

Freeze K &
Restores $W_K$ to its initialization reference every optimizer step\\

QK Decouple &
Begins from equal Q/K initialization, then rotates $W_K$ once after epoch 3 begins\\

QK Align Regularized &
Adds a weak Q/K alignment regularization penalty with $\lambda=0.001$\\

\bottomrule
\end{tabular}

\caption{Controlled interventions.}
\label{tab:controlled-interventions}
\end{table}

Each intervention was evaluated across five random seeds (7, 37, 42, 105, and 210) for 15 epochs using the same geometric metrics and logging framework as the baseline experiments.

\subsection{Token-Level Checkpoint Extension}

To test whether the small character-level results extend to larger token-level language models, we also measured activation-based query/key participation ratios and attention geometry on public intermediate checkpoints. This extension covers GPT-2 small, Pythia 70M/160M/410M, SmolLM2 135M/360M, and OLMo-1B, spanning approximately 70M to 1B parameters. These models use their native tokenizers, corpora, checkpoint schedules, and standard independent query/key initialization; thus the extension is a boundary analysis rather than a controlled replication of the initialization grid. Six checkpoint series contain at least three revisions and support training-progress slope fits; OLMo-1B contributes only an endpoint layer-profile check. Full results are reported in Appendix~\ref{app:scaling}.

\section{Results}

All baseline results are reported across 36 runs spanning four seeds, three depths, and three initialization regimes. Unless otherwise stated, trajectory statistics use seed-averaged initialization-depth-layer traces over the full 45-epoch baseline training horizon, while final geometry and loss summaries average over the corresponding trained runs. Validation losses remain close across initialization regimes: QK Identical reaches $1.1907 \pm 0.0106$, Disjoint Support Xavier reaches $1.1969 \pm 0.0112$, and Independent Xavier reaches $1.1973 \pm 0.0108$. Pairwise Welch tests do not show a large loss separation.\footnote{Pairwise Welch tests: QK Identical vs.\ Independent Xavier, $p=0.144$, Hedges' $g=-0.598$; QK Identical vs.\ Disjoint Support Xavier, $p=0.180$, $g=-0.545$; Independent Xavier vs.\ Disjoint Support Xavier, $p=0.921$, $g=0.039$.} We therefore treat the main effects below as differences in internal attention geometry rather than as simple consequences of different task loss.

\subsection{Persistent Geometry and Layer Roles}

\begin{figure}[H]
\centering
\includegraphics[width=\linewidth]{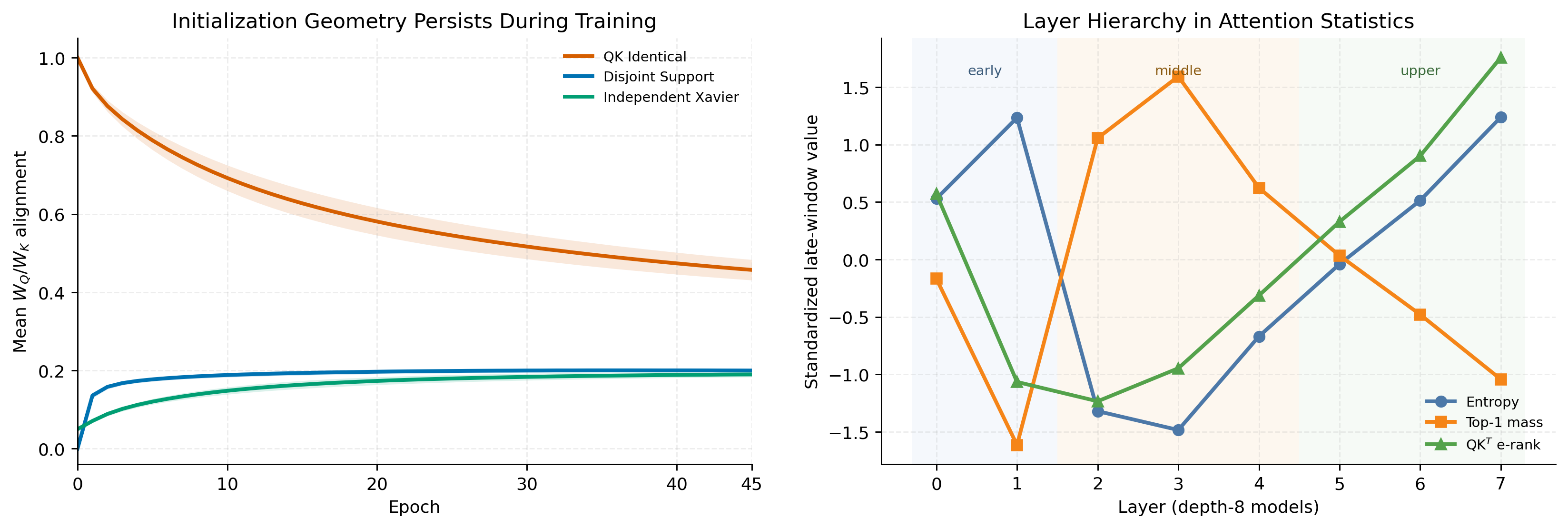}
\caption{Initialization geometry and depth-wise attention roles. Top: $W_Q/W_K$ alignment remains separated by initialization; bottom: depth-8 profiles reveal early, middle, and upper regimes.}
\label{fig:results-geometry-hierarchy}
\end{figure}

The strongest persistent initialization signal is the final alignment between $W_Q$ and $W_K$. QK Identical begins with alignment $1.0000$, while Disjoint Support Xavier begins near $0$ and Independent Xavier begins near $0.0498$. After 45 epochs, the regimes remain separated in the aggregated final summaries: final alignment is $0.4530 \pm 0.0950$ for QK Identical, $0.1991 \pm 0.0151$ for Disjoint Support Xavier, and $0.1895 \pm 0.0227$ for Independent Xavier. The QK Identical versus uncoupled final-alignment contrast remains large at convergence.\footnote{Welch test comparing QK Identical against the combined uncoupled regimes: $p=1.06\times10^{-6}$, Hedges' $g=4.489$.} Thus the initial angular relation is not erased by training. Figure~\ref{fig:results-geometry-hierarchy} and Table~\ref{tab:loss-alignment} summarize this persistent separation.

\begin{table}[!htbp]
\centering
\caption{Final loss and alignment summary.}
\label{tab:loss-alignment}
\resizebox{\columnwidth}{!}{%
\begin{tabular}{lrrrr}
\toprule
Initialization & Runs & Final val loss & Final alignment & Depth-8 alignment \\
\midrule
QK Identical & 12 & $1.1907 \pm 0.0106$ & $0.4530 \pm 0.0950$ & $0.5048 \pm 0.0861$ \\
Disjoint Support Xavier & 12 & $1.1969 \pm 0.0112$ & $0.1991 \pm 0.0151$ & $0.1911 \pm 0.0077$ \\
Independent Xavier & 12 & $1.1973 \pm 0.0108$ & $0.1895 \pm 0.0227$ & $0.1799 \pm 0.0241$ \\
\bottomrule
\end{tabular}
}
\end{table}

The persistence is also depth-dependent. At four layers, the final alignment gap between QK Identical and the strongest uncoupled regime is approximately $0.143$; at eight layers, the same gap grows to approximately $0.314$. Thus depth does not wash out the initialization signature. It amplifies the distance between the coupled and uncoupled geometric regimes while leaving final validation loss comparatively close.

The resulting aggregate final representations can be categorized into three different layer categories. Initial layers have higher entropy and \QKt{} effective rank, which is in line with keeping the token-level information intact. Mid layers make up the most constrained attention bottleneck with average entropy $1.532$, average top-1 mass $0.514$, and average \QKt{} effective rank $6.26$. Higher layers increase the interaction space again as their \QKt{} effective rank increases to $16.34$, while their \PRQ{} becomes $24.41$, contrasted with the middle layer's average value of $10.52$. This confirms the outline-level description of the model training process.

\subsection{Q-Expansion and Key Specialization}

Expansion on the query side grows significantly in the seed-averaged trajectories. This trend is positive for nearly all seed-averaged initialization-depth-layer trajectories and exists in both uncoupled and coupled conditions. As a result, query expansion becomes an established aggregation trend rather than something that is specific to one particular seed or initialization depth.

\begin{figure}[H]
\centering
\includegraphics[width=0.66\linewidth]{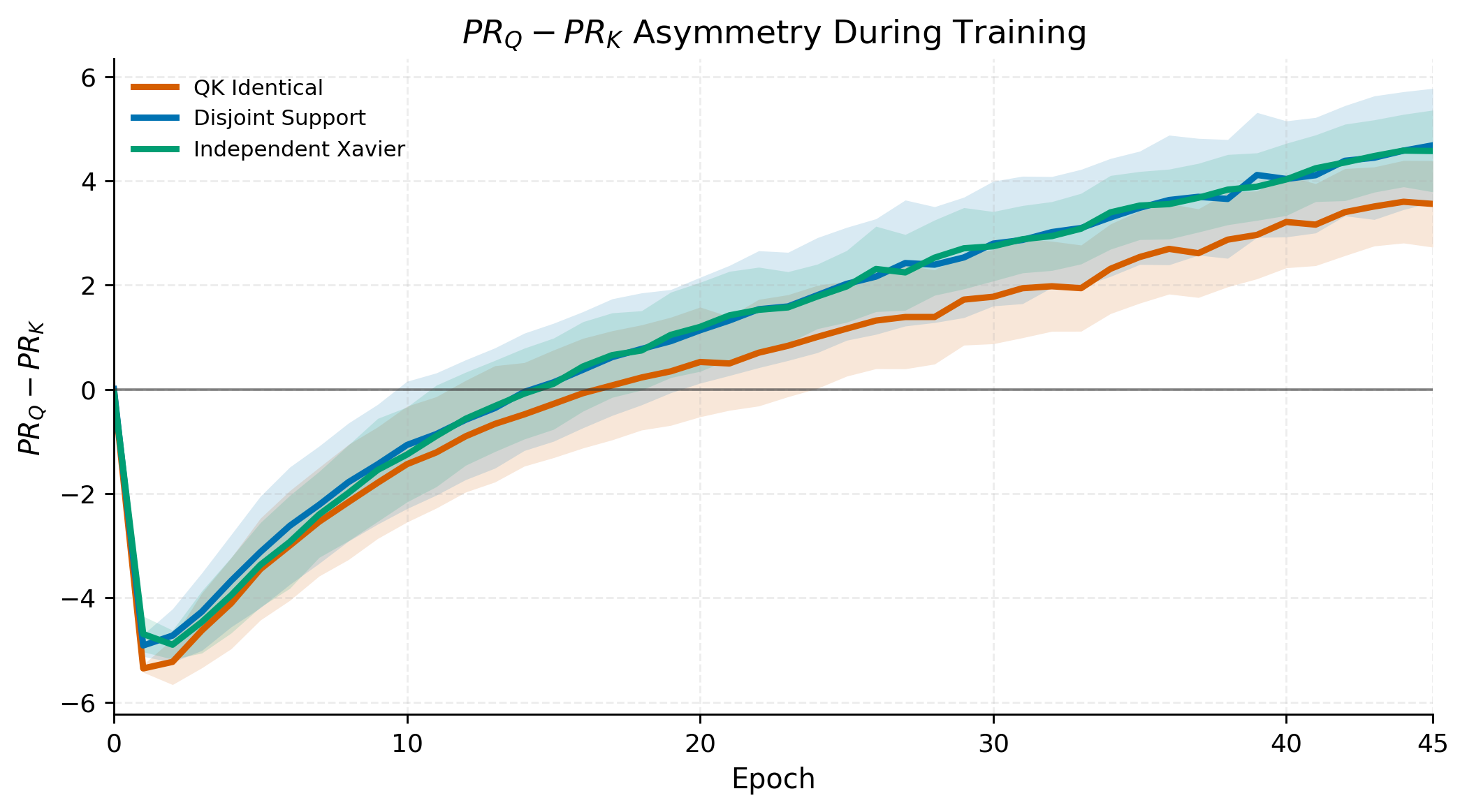}
\caption{Query participation outpaces key participation. Curves show seed-wise averaged $\PRQ-\PRK$ trajectories; shaded bands show standard error across seeds.}
\label{fig:results-participation-summary}
\end{figure}

Participation on the key side demonstrates a complementary specialization trend. Again, in the seed-averaged summary statistics, the slope of \PRK{} is negative for most trajectories and most strongly negative for uncoupled initialization. This means that key-side contraction is markedly weaker under the coupled condition, and reverses sign for a sizable minority of individual trajectories, consistent with the coupled initialization pulling key participation closer to the rising query trajectory. Table~\ref{tab:pr-dynamics} reports these slope contrasts, and Figure~\ref{fig:results-participation-summary} shows the corresponding trajectories.

\begin{table}[!htbp]
\centering
\caption{Participation-ratio slope summary.}
\label{tab:pr-dynamics}
\resizebox{\columnwidth}{!}{%
\begin{tabular}{lrrrr}
\toprule
Subset & $n$ & \PRQ{} slope & \PRK{} slope & $(\PRQ-\PRK)$ slope \\
\midrule
All seed-averaged traces & 54 & $+0.1406 \pm 0.0902$ & $-0.0437 \pm 0.0565$ & $+0.1843 \pm 0.0966$ \\
Uncoupled Q/K & 36 & $+0.1366 \pm 0.0932$ & $-0.0539 \pm 0.0547$ & $+0.1904 \pm 0.0930$ \\
QK Identical & 18 & $+0.1487 \pm 0.0859$ & $-0.0235 \pm 0.0561$ & $+0.1722 \pm 0.1051$ \\
\bottomrule
\end{tabular}
}
\end{table}

The standard deviation on the pooled \PRK{} slope in Table~\ref{tab:pr-dynamics} exceeds its mean, which reflects real heterogeneity in effect size rather than measurement noise: key contraction is strong and consistent under both uncoupled regimes, while under QK Identical it is markedly weaker and occasionally reverses, since the coupled initialization lets the rising query trajectory pull key participation upward at some depths instead of letting it contract. Pooling these regimes into a single mean and standard deviation is therefore expected to look noisy on \PRK{} alone, even though the direction of the effect still holds for the large majority of trajectories: key engagement decreases in 50 of the 54 seed-averaged traces overall.

Paired contrast is the most robust measure precisely because it is far less sensitive to this regime-dependent heterogeneity. The slope of $\PRQ-\PRK$ is positive for every seed averaged initialization depth layer path, whether uncoupled or coupled. It follows that the fundamental asymmetry is neither an expansion of the queries nor a contraction of the keys in isolation, but rather an increasingly broad query path compared to the key path.

\begin{figure}[H]
\centering
\includegraphics[width=\linewidth]{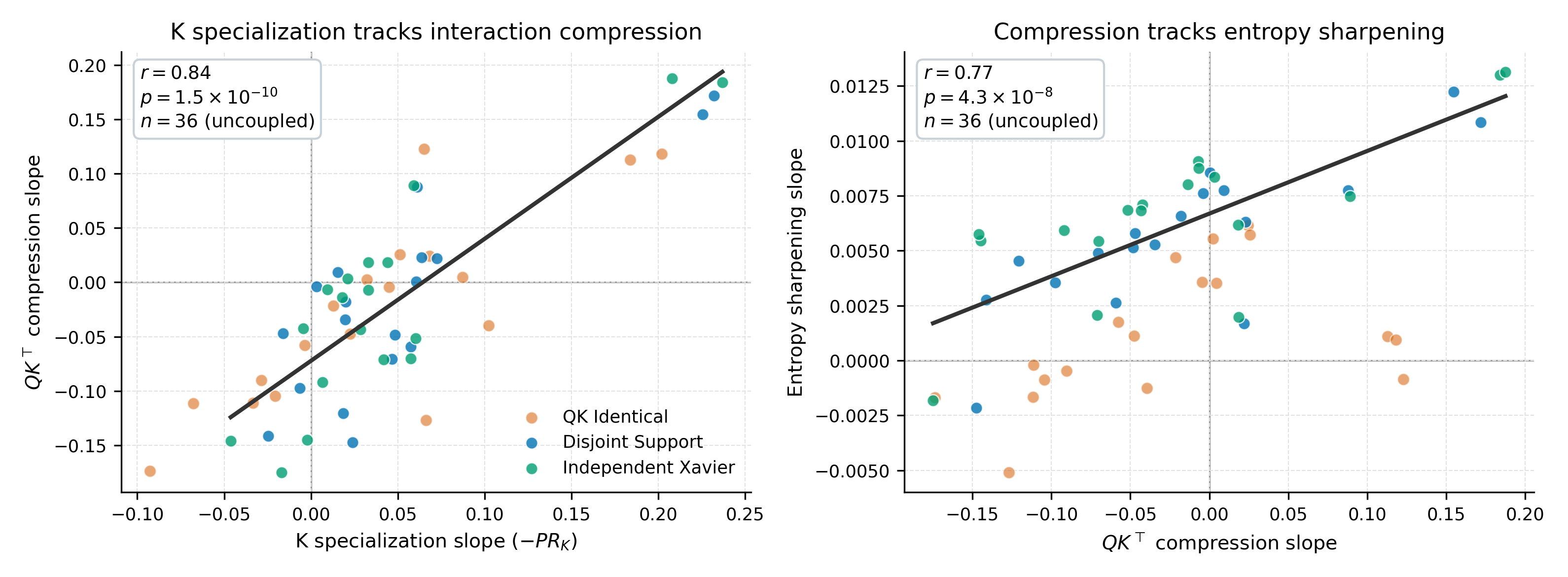}
\caption{Key specialization predicts interaction-spectrum compression. Compression, in turn, predicts entropy sharpening. Each point is one initialization-depth-layer cell with slopes averaged over the four seeds. Fitted lines and statistics use the 36 uncoupled cells; QK Identical cells are shown for reference.}
\label{fig:results-mechanism-diagram}
\end{figure}

Figure~\ref{fig:results-mechanism-diagram} summarizes the measured bridge from key specialization to interaction compression and entropy sharpening.

This relationship is important since participation processes are correlated with both the level of attention sharpness and the structure of interactions-rank. To respect the seed-independence policy of Section~3.3, we first average each run-layer slope over the four seeds within every initialization-depth-layer cell, giving $n=36$ uncoupled observations, the same resolution as the uncoupled row of Table~\ref{tab:pr-dynamics}. At this resolution, higher key specialization is associated with stronger relative compression of the \QKt{} interaction spectrum ($r=0.840$, $p=1.5\times10^{-10}$), and that compression is in turn associated with entropy sharpening ($r=0.769$, $p=4.3\times10^{-8}$) and top-1 sharpening ($r=0.637$, $p=3.0\times10^{-5}$). Treating the four seeds as separate observations ($n=144$) gives the same direction with somewhat weaker correlations ($r=0.781$, $p=8.32\times10^{-31}$; $r=0.614$, $p=2.73\times10^{-16}$; $r=0.488$, $p=5.43\times10^{-10}$), but because those points are not independent we report them only as a descriptive check. Both sets of numbers are computed from the same per-layer slopes, fitted over every logged epoch of all 24 uncoupled baseline runs. Query expansion is therefore best interpreted as a robust background adaptation, while key specialization is the more direct bridge to attention sharpening, with the seed-averaged paired trajectories in Table~\ref{tab:pr-dynamics} providing the independence-respecting confirmation of the underlying asymmetry.

\subsection{Controlled Interventions on K Geometry}

The controlled interventions test whether K-side geometry is merely correlated with attention behavior or can directly modulate it. The observational mechanism predicts that forcing K contraction should sharpen attention by reducing entropy and increasing logit scale, while preserving or freezing the K spectrum should soften attention by maintaining a broader key subspace. The seed-aggregated intervention outcomes in Table~\ref{tab:seed-aggregated-intervention} match this direction: forced K contraction has lower entropy and higher logit standard deviation than the K-preserving interventions, while K-spectrum clamp and freeze K preserve broader interaction spectra.

\begin{table}[!htbp]
\centering
\caption{Seed-aggregated intervention outcomes across seeds 7, 37, 42, 105, and 210. Confidence is the fraction of seed-metric comparisons whose final-epoch delta has the same sign as the aggregate delta for entropy, logit standard deviation, and \QKt{} effective rank.}
\label{tab:seed-aggregated-intervention}
\resizebox{\columnwidth}{!}{%
\begin{tabular}{lrrrrr}
\toprule
Mode & Val loss & Entropy & Logit std & \QKt{} erank & Conf. \\
\midrule
No intervention & 1.2066 & 2.1893 & 4.4329 & 9.71 & -- \\
K spectrum clamp & 1.2159 & 2.2888 & 3.5898 & 11.13 & 1.00 \\
Forced K contraction & 1.2270 & 2.0809 & 5.4071 & 6.03 & 1.00 \\
Freeze Q & 1.2196 & 2.5532 & 3.5915 & 9.35 & 0.93 \\
Freeze K & 1.2172 & 2.6002 & 3.2136 & 10.83 & 1.00 \\
Q/K decouple & 1.2115 & 1.7432 & 5.0723 & 9.92 & 0.93 \\
Q/K alignment regularized & 1.2153 & 2.1767 & 4.6830 & 9.50 & 0.73 \\
\bottomrule
\end{tabular}
}
\end{table}

\begin{figure}[H]
\centering
\includegraphics[width=\linewidth]{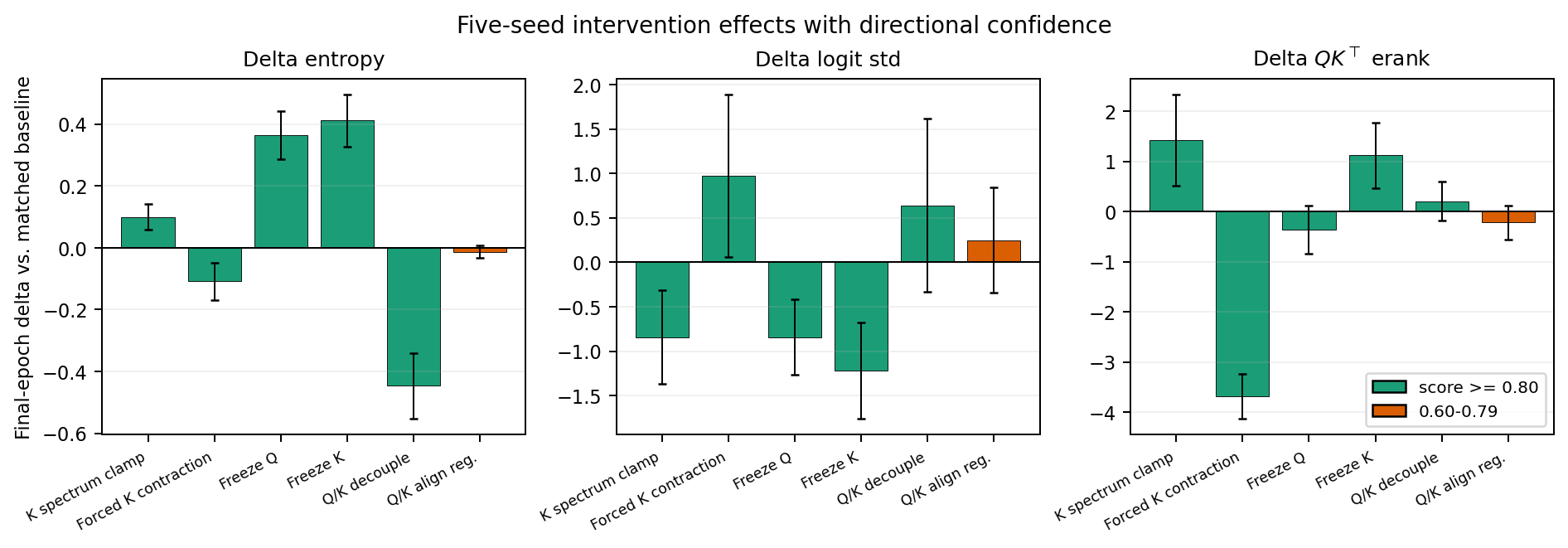}
\caption{Five-seed intervention deltas against each seed's matched no-intervention baseline. Bars show final-epoch mean deltas, error bars show 95\% confidence intervals across seeds, and color indicates directional confidence score.}
\label{fig:results-intervention-confidence}
\end{figure}
\FloatBarrier

The seed-aggregated intervention results show the same qualitative direction across all five seeds. Forced K contraction yields the lowest mean entropy among the independent-initialized intervention conditions ($2.0809$), the largest mean logit standard deviation ($5.4071$), and the lowest \QKt{} effective rank ($6.03$), with confidence score $1.00$. In contrast, freezing K gives mean entropy $2.6002$, mean logit standard deviation $3.2136$, and confidence score $1.00$ in the opposite softening direction. K-spectrum clamp also softens attention relative to the matched no-intervention baseline, with higher entropy, lower logit scale, larger \QKt{} effective rank, and confidence score $1.00$. These effects occur with aggregate final validation losses still in a narrow range from $1.2066$ to $1.2270$ for the independent-initialized intervention family.

The intervention suite also separates K specialization from adjacent explanations. Freezing Q softens attention even though it does not directly freeze the key pathway, indicating that query-side dynamics still affect the final attention state. However, the K-specific manipulations produce the clearest monotonic pattern: contraction sharpens, while preservation or freezing softens. The weak Q/K alignment regularizer produces only modest aggregate changes relative to the stronger K-specific manipulations, which argues against an alignment-only account. These interventions support the interpretation that K geometry acts as a causal control surface for the \QKt{} interaction spectrum and, through it, attention entropy and logit scale.

\subsection{Scaling Behavior on Public Checkpoints}

We also checked whether the small-model pattern carries over to public token-level checkpoints (GPT-2 small, Pythia, and SmolLM2); the full analysis is reported in Appendix~\ref{app:scaling}. In brief, the paired-contrast slope of $\PRQ-\PRK$ is positive in only one of the six fitted checkpoint series, GPT-2 small shows an early $\PRQ-\PRK$ burst that later decays, and the \PRK{}-to-\QKt{}-effective-rank-to-entropy link recurs in four of the six series.

\section{Discussion}

This work aimed to answer the question, how does the effective dimensionality of query and key representations evolve during the training run, and how does initialization affect it. After running the diagnostic experiments, it was observed that the \PRQ{}  increases throughout the training run, while \PRK{}  decreases. The dynamics of participation ratio might be due to complex interaction of initialization, gradients and architecture. We observe that initialization does affect the trajectory of the ratios. 

The increase in \PRQ{} suggests that queries come to occupy a broader activation subspace over training, while the decrease in \PRK{} means keys settle into a narrower one. This is a statement about the spread of the projected query and key activations, not about the rank of the weight matrices $W_q$ and $W_k$ themselves; we track weight-level SVD effective rank separately, and it is not the quantity behind this trend. A broader query space may allow tokens to interact with a larger number of key projections, potentially making the attention more uniform. The contracting key space is consistent with lower entropy from the intervention. Both phenomenon, suggesting that effective attention comes out as a balance between the \PRQ{} and \PRK{}.

Beyond giving an insight into the training process of Transformers, the above-mentioned findings indicate that we can use geometric values like \PRQ{} and \PRK{} for diagnostic purposes to check if the attention is uniformly distributed due to lack of specialization in keys. Such an approach will help in determining the reason behind some unwanted behavior of attention and suitable changes can be made accordingly.

The larger checkpoint extension helps situate this diagnostic within the arc of a full pretraining run. In GPT-2 small, the query-over-key participation gap rises sharply in the first several percent of training and then decays as key participation recovers; the controlled character-level experiments, trained for a comparatively short and fixed horizon, appear to be characterizing this same early regime rather than a property that holds across the whole of pretraining. Under this reading, query expansion and key specialization are best understood as an early-training signature of how attention geometry initially organizes itself, rather than a monotonic law that must hold at every stage of training or at every scale. The \PRK{}-to-\QKt{}-effective-rank-to-entropy bridge discussed above is a comparatively more portable signal across the checkpoint families tested, though it is likewise not a scale-invariant law.

\section{Limitations}

The main experiments are performed on small character-level GPT-like decoder transformers trained on WikiText-103 with context length 128 to permit dense geometric logging with restricted computational resources. The token-level checkpoint extension partially addresses external validity, but it should be interpreted as a boundary analysis rather than a controlled replication: the public checkpoint families differ in tokenizer, corpus, architecture, checkpoint spacing, and training recipe, and they do not permit the same query-key initialization manipulations. Larger controlled token-level experiments, retrieval-focused tasks or longer context length settings, other datasets, longer training schedules, and even transformer variants with grouped/multi-query attention would clarify the extent to which these dynamics persist across different Transformer architectures and variants. In particular, because the GPT-2 checkpoint series shows the query-over-key asymmetry peaking early and then decaying, an open question is whether extending the controlled character-level training horizon well past the current 45 epochs would reveal a similar late-training reversal within the initialization grid itself. As noted above, the interaction-rank-to-entropy bridge is more portable than the raw participation-ratio slope but is not itself scale-invariant, so it should not be over-relied upon as a fallback universal claim.

The participation ratio is a geometric metric of spread or effective dimensionality, not of richness in semantics. This means that the increases in \PRQ{} reflect usage of more spread effective subspace, not richer semantics. Because $\PR(X)$ is invariant to uniform rescaling of the activations, a simple growth in overall activation norm cannot by itself explain the reported \PRQ{}/\PRK{} trends; however, we have not separately verified that direction-dependent (anisotropic) norm growth is not contributing to the measured spectral shifts, and cleanly isolating this from genuine angular reorganization of the query/key subspaces is left as an open question for future work.

Observational evidence is provided for most baselines, however, consistency of the slopes comparison and correlation with cross-metric values provides the clear empirical structure throughout training schedule. Intervention experiments provide additional support for this interpretation and are now balanced across five seeds for the reported 8-layer setting, but they remain short-horizon mechanistic probes. Further ablation over the frequency of interventions, strength of clamp, threshold for ranks, head-wise vs. layer-wise intervention, interactions on the optimizer side are expected. The GPT-2 small extension also combines an early smoke run with a fuller checkpoint sweep, so a clean remeasurement with uniform token budgets would be preferable for stronger scaling claims.

\section{Conclusion}

This work sought to explore whether query and key projections undergo parallel development across initialization, depth, and layers during Transformer training. From 36 baseline runs, query engagement grew in 49 out of 54 seed-based trajectories, while key engagement decreased in 50 out of 54, and the difference $\PRQ-\PRK$ increased in all 54.

The findings also highlight the importance of the discussed asymmetry. It appears that key specialization is more strongly correlated with attention sharpening than query engagement, and that it is connected with the geometry of projections and relative compression of the \QKt{} interaction spectrum. The performed interventions support the causal sensitivity of this pathway: forced compression of the K spectrum reduced entropy and increased logit scale with confidence score $1.00$, whereas maintaining or freezing K spectrum made attention softer with confidence score $1.00$. Token-level checkpoint analyses qualify the scope of the result: monotonic growth of $\PRQ-\PRK$ is not universal across pretrained families, but the link between interaction-rank structure and attention entropy remains the more stable cross-model signal.

The general conclusion from this research is that initialization is not just an optimization starting point but rather a geometric prior on the query-key interface. Attention can be constructed using different internal geometries despite reaching identical validation losses.

\bibliographystyle{plainnat}
\bibliography{references}

\begin{thebibliography}{23}
\providecommand{\natexlab}[1]{#1}
\providecommand{\url}[1]{\texttt{#1}}
\expandafter\ifx\csname urlstyle\endcsname\relax
  \providecommand{\doi}[1]{doi: #1}\else
  \providecommand{\doi}{doi: \begingroup \urlstyle{rm}\Url}\fi

\bibitem[Ataee~Tarzanagh et~al.(2023)Ataee~Tarzanagh, Li, Zhang, and
  Oymak]{tarzanagh2023max}
Davoud Ataee~Tarzanagh, Yingcong Li, Xuechen Zhang, and Samet Oymak.
\newblock Max-margin token selection in attention mechanism.
\newblock In A.~Oh, T.~Naumann, A.~Globerson, K.~Saenko, M.~Hardt, and
  S.~Levine, editors, \emph{Advances in Neural Information Processing Systems},
  volume~36, pages 48314--48362. Curran Associates, Inc., 2023.
\newblock URL
  \url{https://proceedings.neurips.cc/paper_files/paper/2023/file/970f59b22f4c72aec75174aae63c7459-Paper-Conference.pdf}.

\bibitem[Bao et~al.(2024)Bao, Hataya, and Karakida]{bao2024selfattention}
Han Bao, Ryuichiro Hataya, and Ryo Karakida.
\newblock Self-attention networks localize when qk-eigenspectrum concentrates,
  2024.
\newblock URL \url{https://arxiv.org/abs/2402.02098}.

\bibitem[Clark et~al.(2019)Clark, Khandelwal, Levy, and Manning]{clark2019does}
Kevin Clark, Urvashi Khandelwal, Omer Levy, and Christopher~D. Manning.
\newblock What does bert look at? an analysis of bert's attention, 2019.
\newblock URL \url{https://arxiv.org/abs/1906.04341}.

\bibitem[Dong et~al.(2023)Dong, Cordonnier, and Loukas]{dong2023attention}
Yihe Dong, Jean-Baptiste Cordonnier, and Andreas Loukas.
\newblock Attention is not all you need: Pure attention loses rank doubly
  exponentially with depth, 2023.
\newblock URL \url{https://arxiv.org/abs/2103.03404}.

\bibitem[Dong et~al.(2025)Dong, Noci, Khodak, and Li]{dong2025randomattention}
Yihe Dong, Lorenzo Noci, Mikhail Khodak, and Mufan Li.
\newblock Is random attention sufficient for sequence modeling? disentangling
  trainable components in the transformer, 2025.
\newblock URL \url{https://arxiv.org/abs/2506.01115}.

\bibitem[Elhage et~al.(2021)Elhage, Nanda, Olsson, Henighan, Joseph, Mann,
  Askell, Bai, Chen, Conerly, DasSarma, Drain, Ganguli, Hatfield-Dodds,
  Hernandez, Jones, Kernion, Lovitt, Ndousse, Amodei, Brown, Clark, Kaplan,
  McCandlish, and Olah]{elhage2021mathematical}
Nelson Elhage, Neel Nanda, Catherine Olsson, Tom Henighan, Nicholas Joseph, Ben
  Mann, Amanda Askell, Yuntao Bai, Anna Chen, Tom Conerly, Nova DasSarma, Dawn
  Drain, Deep Ganguli, Zac Hatfield-Dodds, Danny Hernandez, Andy Jones, Jackson
  Kernion, Liane Lovitt, Kamal Ndousse, Dario Amodei, Tom Brown, Jack Clark,
  Jared Kaplan, Sam McCandlish, and Chris Olah.
\newblock A mathematical framework for transformer circuits.
\newblock \emph{Transformer Circuits Thread}, 2021.
\newblock https://transformer-circuits.pub/2021/framework/index.html.

\bibitem[Glorot and Bengio(2010)]{glorot2010understanding}
Xavier Glorot and Yoshua Bengio.
\newblock Understanding the difficulty of training deep feedforward neural
  networks.
\newblock In Yee~Whye Teh and Mike Titterington, editors, \emph{Proceedings of
  the Thirteenth International Conference on Artificial Intelligence and
  Statistics}, volume~9 of \emph{Proceedings of Machine Learning Research},
  pages 249--256, Chia Laguna Resort, Sardinia, Italy, 13--15 May 2010. PMLR.
\newblock URL \url{https://proceedings.mlr.press/v9/glorot10a.html}.

\bibitem[Henry et~al.(2020)Henry, Dachapally, Pawar, and
  Chen]{henry2020querykey}
Alex Henry, Prudhvi~Raj Dachapally, Shubham~Shantaram Pawar, and Yuxuan Chen.
\newblock Query-key normalization for transformers.
\newblock In \emph{Findings of EMNLP}, pages 4246--4253, 2020.
\newblock \doi{10.18653/v1/2020.findings-emnlp.379}.
\newblock URL \url{https://aclanthology.org/2020.findings-emnlp.379/}.

\bibitem[Hong and Lee(2025)]{hong2025variance}
Jonghyun Hong and Sungyoon Lee.
\newblock Variance sensitivity induces attention entropy collapse and
  instability in transformers.
\newblock In \emph{EMNLP}, pages 8360--8378, 2025.
\newblock \doi{10.18653/v1/2025.emnlp-main.421}.
\newblock URL \url{https://aclanthology.org/2025.emnlp-main.421/}.

\bibitem[Huang et~al.(2020)Huang, Perez, Ba, and Volkovs]{huang2020improving}
Xiao~Shi Huang, Felipe Perez, Jimmy Ba, and Maksims Volkovs.
\newblock Improving transformer optimization through better initialization.
\newblock In Hal~Daumé III and Aarti Singh, editors, \emph{Proceedings of the
  37th International Conference on Machine Learning}, volume 119 of
  \emph{Proceedings of Machine Learning Research}, pages 4475--4483. PMLR,
  13--18 Jul 2020.
\newblock URL \url{https://proceedings.mlr.press/v119/huang20f.html}.

\bibitem[Jacot et~al.(2020)Jacot, Gabriel, and Hongler]{jacot2020neural}
Arthur Jacot, Franck Gabriel, and Clément Hongler.
\newblock Neural tangent kernel: Convergence and generalization in neural
  networks, 2020.
\newblock URL \url{https://arxiv.org/abs/1806.07572}.

\bibitem[Kobayashi et~al.(2020)Kobayashi, Kuribayashi, Yokoi, and
  Inui]{kobayashi2020attention}
Goro Kobayashi, Tatsuki Kuribayashi, Sho Yokoi, and Kentaro Inui.
\newblock Attention is not only a weight: Analyzing transformers with vector
  norms, 2020.
\newblock URL \url{https://arxiv.org/abs/2004.10102}.

\bibitem[Li et~al.(2026)Li, Tong, Wang, and Hu]{li2026bornbiased}
Siquan Li, Yao Tong, Haonan Wang, and Tianyang Hu.
\newblock Transformers are born biased: Structural inductive biases at random
  initialization and their practical consequences, 2026.
\newblock URL \url{https://arxiv.org/abs/2602.05927}.

\bibitem[Liu et~al.(2020)Liu, Liu, Gao, Chen, and Han]{liu2020understanding}
Liyuan Liu, Xiaodong Liu, Jianfeng Gao, Weizhu Chen, and Jiawei Han.
\newblock Understanding the difficulty of training transformers.
\newblock In Bonnie Webber, Trevor Cohn, Yulan He, and Yang Liu, editors,
  \emph{Proceedings of the 2020 Conference on Empirical Methods in Natural
  Language Processing (EMNLP)}, pages 5747--5763, Online, November 2020.
  Association for Computational Linguistics.
\newblock \doi{10.18653/v1/2020.emnlp-main.463}.
\newblock URL \url{https://aclanthology.org/2020.emnlp-main.463/}.

\bibitem[Noci et~al.(2022)Noci, Anagnostidis, Biggio, Orvieto, Singh, and
  Lucchi]{noci2022signal}
Lorenzo Noci, Sotiris Anagnostidis, Luca Biggio, Antonio Orvieto, Sidak~Pal
  Singh, and Aurelien Lucchi.
\newblock Signal propagation in transformers: Theoretical perspectives and the
  role of rank collapse, 2022.
\newblock URL \url{https://arxiv.org/abs/2206.03126}.

\bibitem[Olsson et~al.(2022)Olsson, Elhage, Nanda, Joseph, DasSarma, Henighan,
  Mann, Askell, Bai, Chen, Conerly, Drain, Ganguli, Hatfield-Dodds, Hernandez,
  Johnston, Jones, Kernion, Lovitt, Ndousse, Amodei, Brown, Clark, Kaplan,
  McCandlish, and Olah]{olsson2022context}
Catherine Olsson, Nelson Elhage, Neel Nanda, Nicholas Joseph, Nova DasSarma,
  Tom Henighan, Ben Mann, Amanda Askell, Yuntao Bai, Anna Chen, Tom Conerly,
  Dawn Drain, Deep Ganguli, Zac Hatfield-Dodds, Danny Hernandez, Scott
  Johnston, Andy Jones, Jackson Kernion, Liane Lovitt, Kamal Ndousse, Dario
  Amodei, Tom Brown, Jack Clark, Jared Kaplan, Sam McCandlish, and Chris Olah.
\newblock In-context learning and induction heads.
\newblock \emph{Transformer Circuits Thread}, 2022.
\newblock
  https://transformer-circuits.pub/2022/in-context-learning-and-induction-heads/index.html.

\bibitem[Pan et~al.(2024)Pan, Philip, Xie, and Schwartz]{pan2024dissecting}
Xu~Pan, Aaron Philip, Ziqian Xie, and Odelia Schwartz.
\newblock Dissecting query-key interaction in vision transformers.
\newblock In A.~Globerson, L.~Mackey, D.~Belgrave, A.~Fan, U.~Paquet,
  J.~Tomczak, and C.~Zhang, editors, \emph{Advances in Neural Information
  Processing Systems}, volume~37, pages 54595--54631. Curran Associates, Inc.,
  2024.
\newblock \doi{10.52202/079017-1730}.
\newblock URL
  \url{https://proceedings.neurips.cc/paper_files/paper/2024/file/6216515a5e0b3257c49dcb1647e497d1-Paper-Conference.pdf}.

\bibitem[Tong et~al.(2026)Tong, Wang, Li, Kawaguchi, and
  Hu]{tong2026seedprints}
Yao Tong, Haonan Wang, Siquan Li, Kenji Kawaguchi, and Tianyang Hu.
\newblock Seedprints: Fingerprints can even tell which seed your large language
  model was trained from, 2026.
\newblock URL \url{https://arxiv.org/abs/2509.26404}.

\bibitem[Vaswani et~al.(2017)Vaswani, Shazeer, Parmar, Uszkoreit, Jones, Gomez,
  Kaiser, and Polosukhin]{vaswani2017attention}
Ashish Vaswani, Noam Shazeer, Niki Parmar, Jakob Uszkoreit, Llion Jones,
  Aidan~N Gomez, \L~ukasz Kaiser, and Illia Polosukhin.
\newblock Attention is all you need.
\newblock In I.~Guyon, U.~Von Luxburg, S.~Bengio, H.~Wallach, R.~Fergus,
  S.~Vishwanathan, and R.~Garnett, editors, \emph{Advances in Neural
  Information Processing Systems}, volume~30. Curran Associates, Inc., 2017.
\newblock URL
  \url{https://proceedings.neurips.cc/paper_files/paper/2017/file/3f5ee243547dee91fbd053c1c4a845aa-Paper.pdf}.

\bibitem[Voita et~al.(2019)Voita, Talbot, Moiseev, Sennrich, and
  Titov]{voita2019analyzing}
Elena Voita, David Talbot, Fedor Moiseev, Rico Sennrich, and Ivan Titov.
\newblock Analyzing multi-head self-attention: Specialized heads do the heavy
  lifting, the rest can be pruned, 2019.
\newblock URL \url{https://arxiv.org/abs/1905.09418}.

\bibitem[Wang et~al.(2022)Wang, Ma, Dong, Huang, Zhang, and
  Wei]{wang2022deepnet}
Hongyu Wang, Shuming Ma, Li~Dong, Shaohan Huang, Dongdong Zhang, and Furu Wei.
\newblock Deepnet: Scaling transformers to 1,000 layers, 2022.
\newblock URL \url{https://arxiv.org/abs/2203.00555}.

\bibitem[Xiong et~al.(2020)Xiong, Yang, He, Zheng, Zheng, Xing, Zhang, Lan,
  Wang, and Liu]{xiong2020layer}
Ruibin Xiong, Yunchang Yang, Di~He, Kai Zheng, Shuxin Zheng, Chen Xing,
  Huishuai Zhang, Yanyan Lan, Liwei Wang, and Tie-Yan Liu.
\newblock On layer normalization in the transformer architecture, 2020.
\newblock URL \url{https://arxiv.org/abs/2002.04745}.

\bibitem[Zhai et~al.(2023)Zhai, Likhomanenko, Littwin, Busbridge, Ramapuram,
  Zhang, Gu, and Susskind]{zhai2023stabilizing}
Shuangfei Zhai, Tatiana Likhomanenko, Etai Littwin, Dan Busbridge, Jason
  Ramapuram, Yizhe Zhang, Jiatao Gu, and Joshua~M. Susskind.
\newblock Stabilizing transformer training by preventing attention entropy
  collapse.
\newblock In Andreas Krause, Emma Brunskill, Kyunghyun Cho, Barbara Engelhardt,
  Sivan Sabato, and Jonathan Scarlett, editors, \emph{Proceedings of the 40th
  International Conference on Machine Learning}, volume 202 of
  \emph{Proceedings of Machine Learning Research}, pages 40770--40803. PMLR,
  23--29 Jul 2023.
\newblock URL \url{https://proceedings.mlr.press/v202/zhai23a.html}.

\end{thebibliography}

\appendix

\section{Scaling Behavior on Public Checkpoints}
\label{app:scaling}

Here we report the token-level checkpoint analysis described in the Experimental Setup. Unlike the main experiments, these models were not trained by us, so we could not vary their initialization. The question is simply whether the pattern we see in small character-level models also shows up in public pretrained checkpoints.

GPT-2 small is the most useful case, because it is the only series with dense checkpoints across nearly all of training. Early on it behaves much like our small models, with $\PRQ-\PRK$ rising sharply within the first few percent of training. After that, key participation recovers and the gap shrinks, so the slope fitted over the whole run is negative ($-0.1098$ after dividing by head dimension), and no individual GPT-2 layer has a positive slope. Our character-level models are trained for a short, fixed horizon, so we think they capture the early phase that GPT-2 small goes through and then leaves, not a trend that continues for all of pretraining.

Table~\ref{tab:token-checkpoint-extension} adds five more series, namely Pythia 70M, 160M and 410M and SmolLM2 135M and 360M, each with at least three checkpoints. OLMo-1B had too few step-tagged checkpoints to fit a slope, so we only used its final checkpoint. Across the six series we could fit, the $\PRQ-\PRK$ slope is positive in just one model (Pythia-70M) and in 17 of 116 individual layers. Over full pretraining, then, the steady growth in $\PRQ-\PRK$ from our controlled grid does not hold in general.

\begin{table}[!htbp]
\centering
\caption{Token-level checkpoint extension. Counts use only checkpoint series with at least three revisions. Bridge co-movement indicates positive correlation between \PRK{} and \QKt{} effective rank together with positive correlation between \QKt{} effective rank and attention entropy.}
\label{tab:token-checkpoint-extension}
\resizebox{\columnwidth}{!}{%
\begin{tabular}{lrrrrr}
\toprule
Family & Models & $\Delta PR$ slope $>0$ & \PRQ{} slope $>0$ & \PRK{} slope $<0$ & Bridge co-move \\
\midrule
Pythia & 3 & 1/3 & 0/3 & 3/3 & 3/3 \\
SmolLM2 & 2 & 0/2 & 0/2 & 0/2 & 0/2 \\
GPT-2 & 1 & 0/1 & 1/1 & 0/1 & 1/1 \\
\midrule
All fitted series & 6 & 1/6 & 1/6 & 3/6 & 4/6 \\
\bottomrule
\end{tabular}
}
\end{table}

\begin{figure}[H]
\centering
\includegraphics[width=\linewidth]{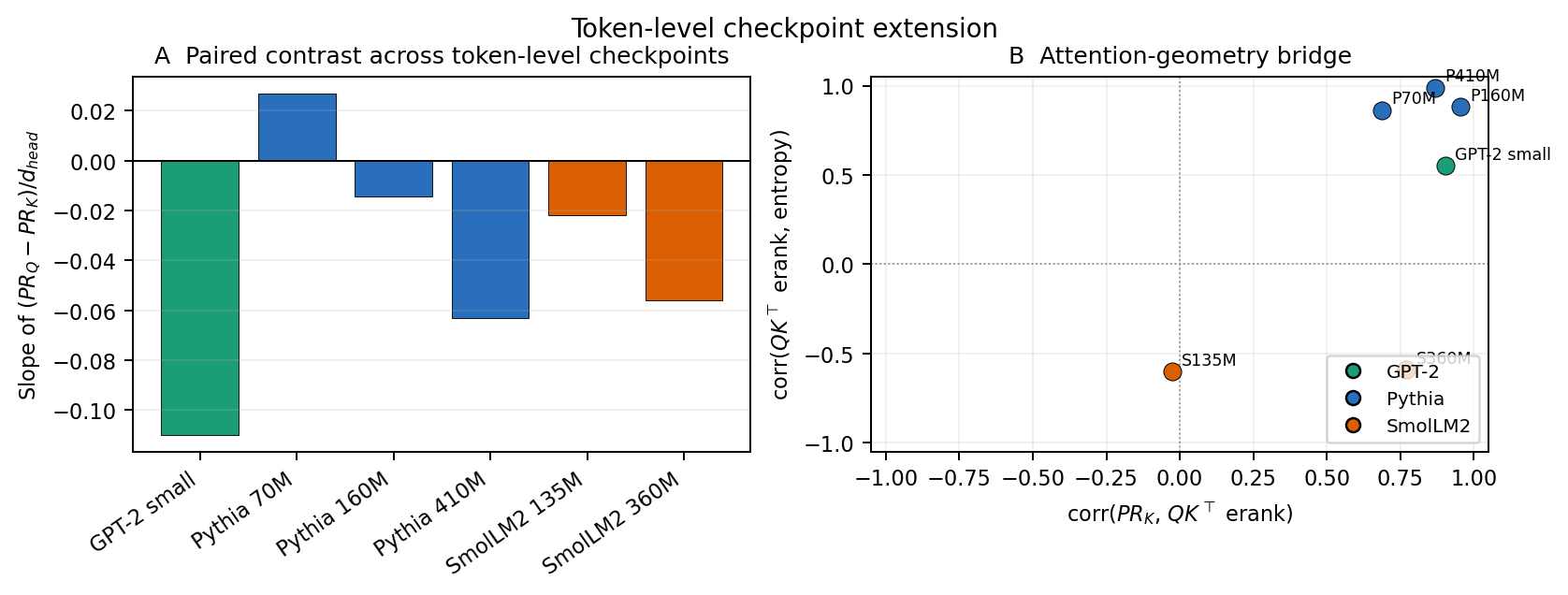}
\caption{Token-level checkpoint extension. The left panel shows fitted slopes of the normalized paired contrast $(\PRQ-\PRK)/d_{head}$ across public checkpoint series. The right panel shows model-level correlations connecting key participation, \QKt{} effective rank, and attention entropy.}
\label{fig:results-token-checkpoint-extension}
\end{figure}
\FloatBarrier

The link between key geometry and attention holds up better than the slope itself. In GPT-2 small, \PRK{} tracks \QKt{} effective rank closely ($r=0.905$), and \QKt{} effective rank in turn tracks attention entropy ($r=0.551$). All three Pythia models show the same pattern, and key participation contracts in each of them. The SmolLM2 models are the exception, since neither shows key contraction or this link. So the connection from \PRK{} through \QKt{} effective rank to entropy appears in four of the six series. It recurs often, but not in every model family we tested.

\end{document}